\documentclass[conference]{IEEEtran}
\usepackage{lineno,hyperref}
\usepackage{multirow}
\usepackage{graphicx}
\usepackage{amsmath}
\usepackage{amssymb}

\ifCLASSINFOpdf
 
\begin{document}

\title{Impact of ultrasound image reconstruction method on breast lesion classification with neural transfer learning}

\author{\IEEEauthorblockN{Micha\l{} Byra\IEEEauthorrefmark{1}\IEEEauthorrefmark{2}\IEEEauthorrefmark{4},
Tomasz Sznajder\IEEEauthorrefmark{3},
Danijel Kor\v{z}inek\IEEEauthorrefmark{3},\\ 
Hanna Piotrzkowska-Wr\'oblewska\IEEEauthorrefmark{4},
Katarzyna Dobruch-Sobczak\IEEEauthorrefmark{4}\IEEEauthorrefmark{5},
Andrzej Nowicki\IEEEauthorrefmark{4} and
Krzysztof Marasek\IEEEauthorrefmark{3}}
\IEEEauthorblockA{\IEEEauthorrefmark{2}Department of Radiology, University of California, San Diego, USA}
\IEEEauthorblockA{\IEEEauthorrefmark{4}Department of Ultrasound, Institute of Fundamental Technological Research, \\Polish Academy of Sciences, Warsaw, Poland}
\IEEEauthorblockA{\IEEEauthorrefmark{3}Department of Multimedia, Polish-Japanese Academy of Information Technology, Warsaw, Poland}
\IEEEauthorblockA{\IEEEauthorrefmark{5}The Maria Sk\l{}odowska-Curie Memorial Cancer Centre and Institute of Oncology, Warsaw, Poland}
\IEEEauthorblockA{\IEEEauthorrefmark{1}Corresponding author, e-mail: mbyra@ucsd.edu}
}

\maketitle

\begin{abstract}

Deep learning algorithms, especially convolutional neural networks, have become a methodology of choice in medical image analysis. However, recent studies in computer vision show that even a small modification of input image intensities may cause a deep learning model to classify the image differently. In medical imaging, the distribution of image intensities is related to applied image reconstruction algorithm. In this paper we investigate the impact of ultrasound image reconstruction method on breast lesion classification with neural transfer learning. Due to high dynamic range raw ultrasonic signals are commonly compressed in order to reconstruct B-mode images. Based on raw data acquired from breast lesions, we reconstruct B-mode images using different compression levels. Next, transfer learning is applied for classification. Differently reconstructed images are employed for training and evaluation. We show that the modification of the reconstruction algorithm leads to decrease of classification performance. As a remedy, we propose a method of data augmentation. We show that the augmentation of the training set with differently reconstructed B-mode images leads to a more robust and efficient classification. Our study suggests that it is important to take into account image reconstruction algorithms implemented in medical scanners during development of computer aided diagnosis systems.
 
\end{abstract}


\IEEEpeerreviewmaketitle

\section{Introduction}

Ultrasound (US) imaging is widely used for breast lesion detection and differentiation. In comparison to other imaging modalities, such as magnetic resonance imaging, mammography or thomosynthesis, US is safe, accurate low cost and highly universalize. These features make the US imaging suitable for breast cancer screening, lesion monitoring and preoperative staging. However, US imaging is operator dependent. Data acquisition needs to be carried out by radiologists or physicians who know how to efficiently operate an ultrasound scanner. The operator has to locate the lesion within the examined breast and properly record the US images. Moreover, interpretation of the US images is not straightforward, but requires deep knowledge of characteristic image features connected to specific lesion types.

So far, various computer-aided diagnosis (CADx) systems have been proposed to support the radiologists and to improve differentiation of malignant and benign breast lesions \cite{Cheng2010299,giger2013breast}. These systems usually use hand-crafted features for classification. Although good classification performance was reported \cite{GomezFlores20151125}, with the current rise of deep learning (DL) methods CADx systems with automatic feature extraction are gaining momentum for breast lesion differentiation \cite{antropova2017deep, han2017deep, litjens2017survey, yap2017automated}. DL methods commonly apply convolutional neural networks to process US images in order to provide the probability that the examined breast lesion is malignant. However, deep neural networks require large datasets for efficient training. This requirement is rarely met in the case of US imaging. Due to this issue, researchers commonly rely on transfer learning methods for CADx system development \cite{shin2016deep}. The aim of the transfer learning is to employ a DL model pre-trained on an existing, large dataset from a different domain to address the problem of interest. This gives the advantage that instead of having to handcraft features, the pre-trained DL model is used to extract the features for classification. Performance of the transfer learning relies on the similarity of the medical images to the images that were used to develop the pre-trained DL model. 

In this paper we investigate the impact of US image reconstruction algorithm on transfer learning based classification of breast lesions. Our study is motivated by several papers that report unstable properties of DL models. Notably because even a small modification of image intensities may cause a well performing DL model to drastically change the output \cite{nguyen2015deep, su2017one}. The visibility of tissue in US images is related to applied image reconstruction techniques. Modification of the reconstruction algorithm properties results in change of US image intensities, which may have a negative impact on DL based classification. We investigate this issue using raw US data acquired from breast lesions. First, raw data are used to reconstruct US images. Next, a classifier is developed using transfer learning. The classification performance is evaluated on two test sets. The first one contains images reconstructed in the same way as in the training set. The second set contains images reconstructed differently. Several experiments are performed. Results and a discussion are presented at the end of the paper.

\section{Materials and methods}

\subsection{Dataset}

In this study we used an extension of the freely available breast lesion dataset, the OASBUD (Open Access Series of Breast Ultrasonic Data) \cite{piotrzkowska2017open}. The OASBUD includes raw US data (before image reconstruction) recorded from breast focal lesions during routine scanning performed in the Maria Sk\l{}odowska Curie Memorial Cancer Centre and Institute of Oncology in Warsaw. The dataset was designed to study quantitative ultrasound techniques \cite{byra2016classification, dobruch2017usefulness}. In comparison to the original dataset that contains raw data acquired from 100 patients, the extended dataset contains data from 251 tumors that were categorized as BI-RADS 3, 4 or 5 . The dataset includes raw data matrices from 94 and 157 malignant and benign lesions, respectively. The pathology of lesions categorized as BI-RADS 4 and 5 was proven by a core needle biopsy. BI-RADS 3 lesions were assessed using fine needle aspiration or deemed benign upon examination and a two year observation (every six months). For each lesion, two orthogonal scans were acquired. For each scan a region of interest (ROI) was determined by an experienced radiologist to correctly indicate lesion area. This study was approved by the Institutional Review Board. More details regarding the dataset can be found in the original paper \cite{piotrzkowska2017open}.

\subsection{Preprocessing}

\begin{figure}[t]
	\begin{center}
		\includegraphics[width=.7\linewidth]{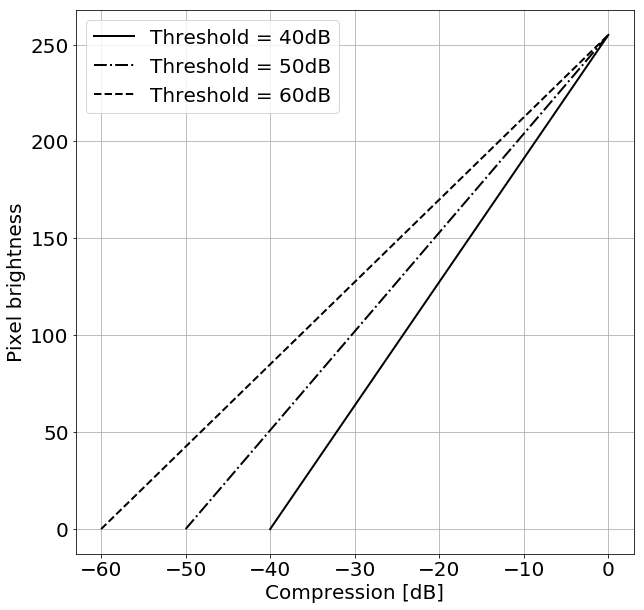}
	\end{center}
	\caption{Mapping function for logarithmic compression using different threshold levels}
	\label{fig:comp}
\end{figure}

\begin{figure}[t]
	\begin{center}
		\includegraphics[width=0.7\linewidth]{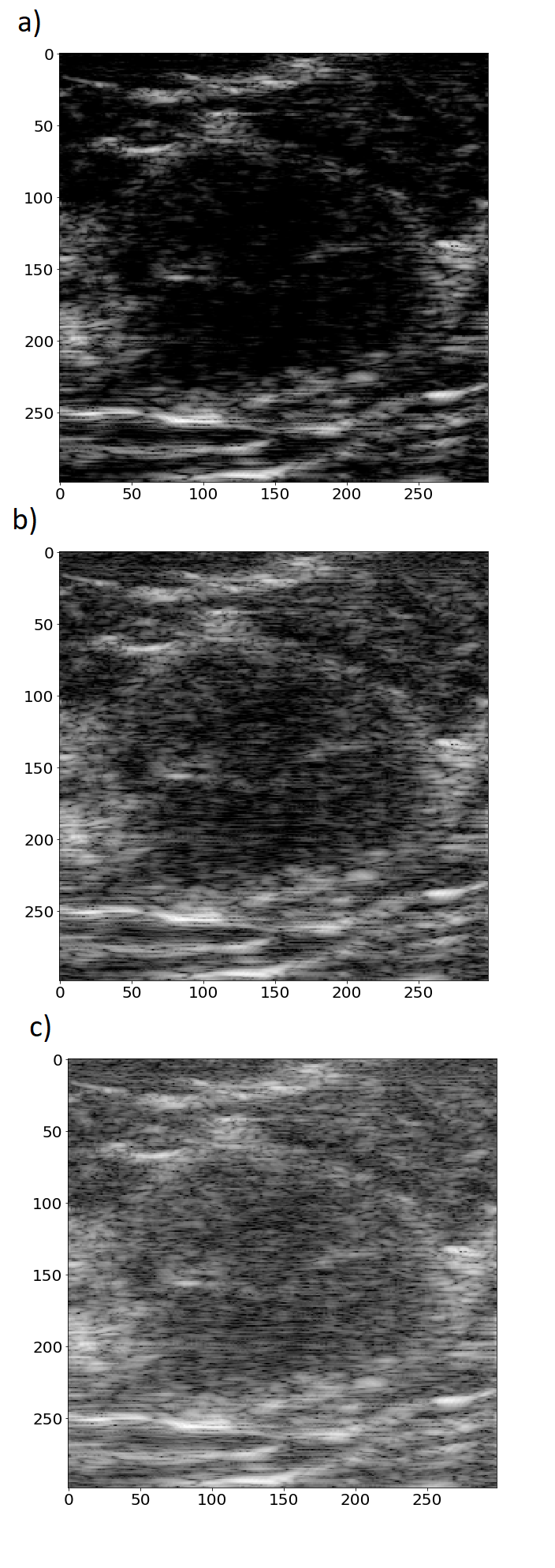}
	\end{center}
	\caption{B-mode images of a benign lesion, applied compression threshold levels of a) 40dB, b) 50dB and c) 60dB. }
	\label{fig:razem}
\end{figure}

\begin{figure*}[t]
	\centering
		\includegraphics[width=0.9\linewidth]{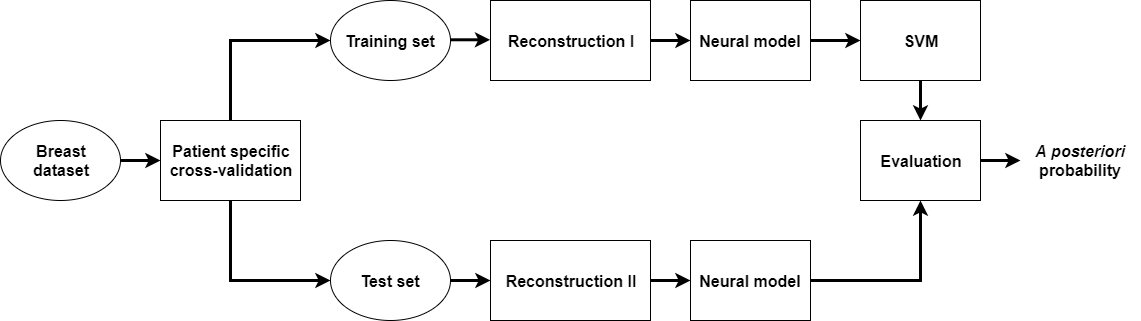}
	\caption{Evaluation pipeline.}
	\label{fig:pipeline}
    \hfil
\end{figure*}

The dynamic range of raw US signals is too high to fit on the screen directly. Therefore, US signals are commonly compressed. To reconstruct the B-mode images we applied the following procedure. First, the envelope of each raw US signal was calculated by the Hilbert transform. Next, the envelope was log-compressed using the following formula: 
\begin{equation}
A_{log} = 20\text{log}_{10}(A/A_{max})
\end{equation}
\noindent where $A$ and $A_{log}$ are the amplitude and the log-compressed amplitude of the ultrasonic signal, respectively. $A_{max}$ indicates the highest value of the amplitude in the data. Next, a specific threshold level was selected and the log-compressed amplitude was mapped to the range of [0, 255]. Three different threshold levels commonly used in US imaging for image reconstruction were applied: 40 dB, 50 dB and 60 dB. Fig. \ref{fig:comp} shows the mapping functions for these different threshold levels. Physicians commonly select the threshold level to obtain the required image quality e.g. good speckle pattern visibility or edge enhancement. Fig. \ref{fig:razem} presents US images of the same lesion reconstructed using different threshold levels. For example, setting the threshold level to 40 dB results in removal of speckles that originates from US echoes of low intensities. Setting this level incorrectly may cause either introduction of too much noise or the removal of important detail in the resulting images.

\subsection{Neural networks}

In our study, we employed two deep CNNs for transfer learning, namely the InceptionV3 \cite{szegedy2016rethinking} and the VGG19 \cite{simonyan2014very}. Both neural networks were pre-trained on the ImageNet dataset \cite{deng2009imagenet} and implemented in TensorFlow \cite{abadi2016tensorflow}. These models achieve state-of-the-art performance on the ImageNet dataset and were recently used for breast lesion classification \cite{antropova2017deep, han2017deep}.

\subsection{Transfer learning and evaluation}

Two experiments were performed and in both cases the classification was evaluated by patient specific stratified 5-fold cross-validation. CNNs were evaluated separately. In the first experiment, to evaluate the impact of reconstruction on the classification performance, the following procedure was applied. For patients in the training set, the images of their lesions were reconstructed using a specified threshold level. Those images were used to develop a classifier. Next, the classifier was evaluated on two test sets. In the first case, the images were reconstructed in the same way as in the case of images in the training set. In the second case, the images were reconstructed using a different threshold level, see Fig. \ref{fig:pipeline}. In the second experiment, images in the training set were reconstructed using different threshold levels and the testing procedure was the same as in the first experiment.

Two methods of data augmentation were applied in our study. The first one was related to the reconstruction of the US images using different threshold levels and was described in the previous section. The second one was related to the fact that breast lesions usually vary in size. The size of a lesion depends on several factors including disease progression and imaging plane (US images are 2D projections of 3D objects). To resize the US images we applied the following procedure. Each US image was cropped using the ROI provided by the radiologist to contain the lesion and a 2, 5 or 10 mm band of surrounding tissue. Next, the US images were resized using bi-cubic interpolation to match the resolution originally designed for each neural network. Intensities of each image were copied along RGB channels and preprocessed in the same way as in the original papers \cite{szegedy2016rethinking, simonyan2014very}. In the case of the InceptionV3 model, features for classification were extracted using the last average pooling layer. In the case of the VGG19 model, the first fully connected layer was used. We used this transfer learning technique because it is general. Moreover, this approach corresponds to the method applied in \cite{antropova2017deep} for breast lesion classification in ultrasound.

We could distinguish three sets of US images that correspond to different threshold levels. For example, one set was related to the threshold level of 40 dB. It contained six US images of the same lesion that originate from two orthogonal scans resized using three different scalings. By $\text{Train}_{40\text{dB}}$ we indicated that the training set contained images reconstructed using threshold level of 40 dB. Moreover, $\text{Train}_{\text{ALL}}$ was related to the training set that contained US images reconstructed using all three threshold levels. 

The support vector machine (SVM) algorithm was employed for classification \cite{chang2011libsvm}. Across all training tests the same SVM classifier with linear kernel was used. Hyper-parameters $C$ and $\gamma$ were set to 1 and 0.001, respectively. To address the problem of class imbalance, we used class weights inversely proportional to class frequencies in the training set. To determine the \textit{a posteriori} probability that a lesion in the test set is malignant, we averaged the probabilities obtained for all images corresponding to this lesion.

The area under (AUC) the receiver operating curve (ROC) was used to assess classification performance. The sensitivity, specificity and accuracy of the classifiers were calculated based on the ROC curve for the point on the curve that was the closest to (0, 1) \cite{fawcett2006introduction}. All calculations were performed in Python. 

\section{Results}

\begin{figure}[t]
	\begin{center}
		\includegraphics[width=.8\linewidth]{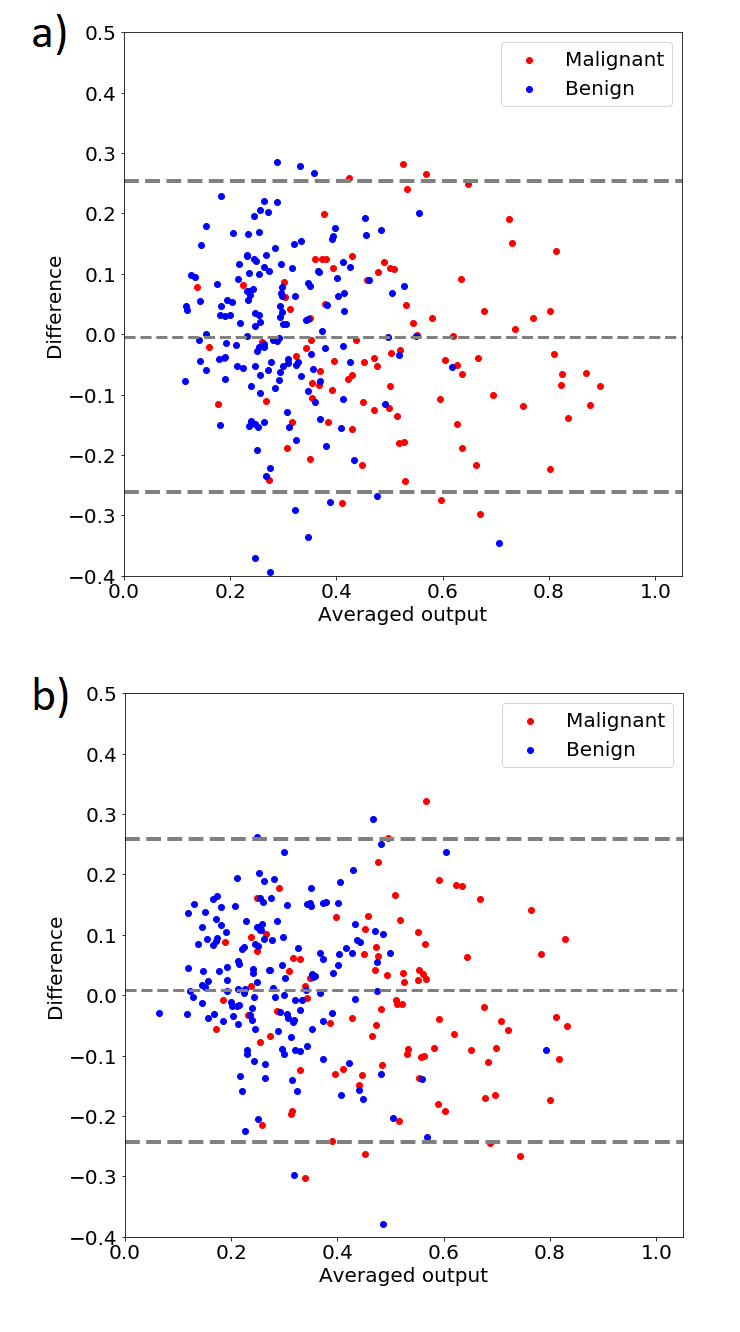}
	\end{center}
	\caption{The Bland-Altman plots illustrating  agreement (estimated probability of malignancy) between the classifiers developed using images reconstructed at compression level of 40 dB and 60 dB. Features were extracted using a) the InceptionV3 and b) the VGG19.}
	\label{fig:altman}
\end{figure}

Tables I shows the classification performance of the classifiers developed using InceptionV3 model. The classifiers were trained and evaluated using differently reconstructed images. Table II shows similar results obtained for the VGG19 model. Tables depict that the approach utilizing InceptionV3 model achieves overall better classification performance. Moreover, the higher is the difference in the threshold levels between the training and the test set, the lower is the classification performance. In the worst case, the AUC value decreased by around 0.1 which corresponded to a drop in accuracy by about 10\%. The Bland-Altman plots in Fig. \ref{fig:altman} illustrate the agreement between the classifiers developed using images reconstructed at compression level of 40 dB and 60 dB. 

As we can see in Tables I and II, using $\text{Train}_{\text{ALL}}$ for training leads to a more robust classification. In both cases, the proposed augmentation technique improved the results, the classification performance across all test sets increased and was less sensitive to modification of the compression threshold level. 

\begin{figure}[t]
	\begin{center}
		\includegraphics[width=.7\linewidth]{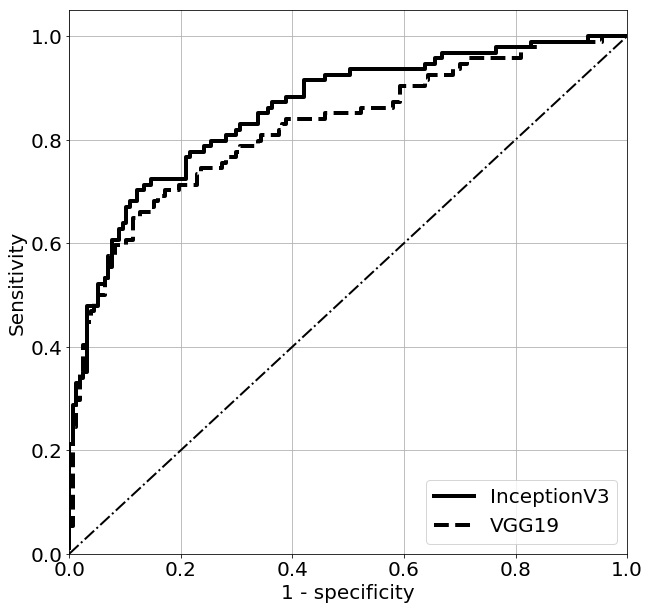}
	\end{center}
	\caption{The ROC curves for the SVM classifiers developed using $\text{Train}_{\text{ALL}}$ and evaluated on $\text{Test}_{\text{ALL}}$, InceptionV3: AUC=0.857, VGG19: AUC=0.822.}
	\label{fig:roc}
\end{figure}

Table III depicts the classification performance of the classifiers developed using $\text{Train}_{\text{ALL}}$ and evaluated on $\text{Test}_{\text{ALL}}$. In the case of the InceptionV3 model, we obtained the best classification performance with the AUC value equal to 0.857. The AUC value for the approach utilizing the VGG19 model was equal to 0.822. Fig. \ref{fig:roc} shows the ROC curves obtained for these best performing classifiers.

\begin{table}[b]
	\begin{center}
		\begin{tabular}{|c|c|c|c|}
            \hline
			  & $\text{Test}_{40\text{dB}}$ & $\text{Test}_{50\text{dB}}$ & $\text{Test}_{60\text{dB}}$ \\
             \hline
			 $\text{Train}_{40\text{dB}}$ & 0.843 & 0.772 & 0.729\\
             \hline
			 $\text{Train}_{50\text{dB}}$ & 0.731  & 0.813 & 0.811\\
			 \hline
			 $\text{Train}_{60\text{dB}}$ & 0.711  & 0.759 & 0.801\\
             \hline \hline
             $\text{Train}_{\text{ALL}}$ & 0.854 & 0.826 & 0.828\\
			\hline
		\end{tabular}
	\end{center}
	\caption{Classification performance of the SVM classifiers developed and evaluated using the InceptionV3 network and differently reconstructed US images.}
	\label{tab:aucvsauc}
\end{table}

\begin{table}[b]
	\begin{center}
		\begin{tabular}{|c|c|c|c|}
            \hline
			  & $\text{Test}_{40\text{dB}}$ & $\text{Test}_{50\text{dB}}$ & $\text{Test}_{60\text{dB}}$ \\
             \hline
			 $\text{Train}_{40\text{dB}}$ & 0.844 & 0.762 & 0.752\\
             \hline
			 $\text{Train}_{50\text{dB}}$ & 0.636  & 0.757 & 0.756\\
			 \hline
			 $\text{Train}_{60\text{dB}}$ & 0.701  & 0.726 & 0.791\\
             \hline \hline
             $\text{Train}_{\text{ALL}}$ & 0.824 & 0.773 & 0.799\\
			\hline
		\end{tabular}
	\end{center}
	\caption{Classification performance of the SVM classifiers developed and evaluated using the VGG19 network and differently reconstructed US images.}
	\label{tab:aucvsauc_vgg}
\end{table}

\begin{table}[t]
	\begin{center}
		\begin{tabular}{|c|c|c|c|c|c|}
			\hline
			 CNN & AUC & Accuracy & Sensitivity & Specificity \\
			\hline
			 InceptionV3 & 0.857 & 0.781 & 0.777 & 0.783 \\
             \hline
			 VGG19 & 0.822 & 0.828 & 0.702 & 0.781 \\
			\hline
		\end{tabular} 
	\end{center}
	\caption{Classification performance of the SVM classifiers developed using $\text{Train}_{\text{ALL}}$ and evaluated on $\text{Test}_{\text{ALL}}$.}
	\label{tab:param}
\end{table}

\section{Discussion}

According to Tables I and II, the classification performance decreases if the test set contains images reconstructed differently  than in the training set. Modification of US image intensities affects feature extraction and has negative impact on the breast lesion classification. Both neural models could not provide features that would be invariant to small modifications of input images' intensities. Accurately recognized images were mislabeled after the modification. Moreover, the Bland-Altman plots in Fig. \ref{fig:altman} illustrate the disagreement between the classifiers developed using 40dB and 60. The difference in the threshold levels was the highest in this case and caused the highest disagreement, see Table I and II. The classifiers developed using $\text{Train}_{40\text{dB}}$ achieved better classification performance. This might be explained by the fact that high image compression is related to edge enhancement, see Fig. \ref{fig:razem}. As reported in several studies, more irregular lesion boundary is expected in the case of malignant lesions \cite{chen2003breast, chou2001stepwise}. Moreover, morphological features are considered to be the most effective handcrafted features for breast lesion classification \cite{GomezFlores20151125}. 

Tables I and II depicts that using $\text{Train}_{\text{ALL}}$ leads to a more robust classification. The proposed technique of data augmentation that employs images reconstructed using different threshold levels leads to better classification performance. In some sense, the proposed method is similar to Gaussian blurring and other techniques that modify image intensities. However, it is specific for the US imaging. 

Our study shows the usefulness of the transfer learning for breast lesion classification. According to Table III, the best overall classification performance was obtained for the classifier developed using $\text{Train}_{\text{ALL}}$ and the InceptionV3 model, with AUC of 0.857. In the case of the VGG19 neural network, the AUC value was equal to 0.822. In \cite{antropova2017deep} the VGG19 neural network and the same transfer learning method as ours was used to classify breast lesions in ultrasound. The authors reported a little bit higher AUC value of 0.85. In \cite{han2017deep} a specific approach to transfer learning was applied, which included modification of the InceptionV3 architecture and the ImageNet dataset. The authors used an ensemble of DL models for classification and reported high AUC value of 0.9601. In our case, we used the InceptionV3 model for transfer learning in a more standard following the approach proposed in \cite{antropova2017deep} what makes direct comparison of the results difficult. 

Our results depict several issues related to neural transfer learning. First of all, the image reconstruction procedures implemented in medical scanners should be taken into account during CADx system development. It is important to know how the medical images were acquired and reconstructed. In our study we used a unique dataset including raw ultrasound signals acquired with a research US scanner. However, usually little is known about the image reconstruction algorithms implemented in US scanners. Usually the researchers involved in CADx systems development agree that a particular system might not test well on data acquired in another medical center using different scanners and protocols. Our study clearly shows that this issue might be related also to a CADx system developed using data recorded in the same medical center. Another problem is connected to US physics. Due to wave attenuation, US echoes coming from larger depths have lower intensities and therefore are differently mapped to image intensity levels. The same object may look slightly differently depending on its depth. Moreover, in this study we examined only the impact of compression threshold level on classification, but other factors exist that can decrease the performance. For example, the texture of the image might depend on applied beamforming technique \cite{yu2015beamforming} or imaging frequency \cite{tsui2017effect}. The proposed method of data augmentation method is general and can be applied to all kinds of US images to improve the CADx system performance with deep learning. However, the proposed method requires raw ultrasound data what limits its usefulness. Nevertheless, our study suggests how to improve the classification performance and the robustness. First, several B-mode images of the same tissue should be acquired with different scanner settings. Second possibility is to always use the same image reconstruction algorithms and scanner settings. This requirement, however, might be difficult to fulfill in reality. One more possibility is to use quantitative ultrasound techniques. These method can be used to create scanner independent parametric maps that illustrate various physical properties of tissue \cite{byra2016classification, sadeghi2017breast}. 

\section{Conclusions}

In this paper we investigated the impact of US image reconstruction algorithm on breast lesion classification with transfer learning. Modification of the reconstruction algorithm leads to decrease of classification performance. To minimize this effect, we proposed a method of data augmentation. We presented that by using differently reconstructed US images for training a better classification performance can be obtained. We believe that our approach is general and can be used to improve the performance of any CADx system that utilizes US images and convolutional neural networks.  

\section*{Acknowledgement} 

This work was partially supported by the National Science Center (Poland), Grant Number UMO-2014/13/B/ST7/01271.

\section*{Conflict of interest statement}

The authors do not have any conflict of interest. 

\bibliographystyle{IEEEtran}
\bibliography{mybibfile}

\end{document}